\documentclass{article}

\usepackage[numbers]{natbib}
\usepackage[preprint]{neurips_2024}
\usepackage{graphicx}
\usepackage{amsmath} 
\usepackage{caption}
\usepackage{float}
\usepackage{booktabs}
\usepackage{array}
\usepackage{placeins}

\usepackage[utf8]{inputenc} % allow utf-8 input
\usepackage[T1]{fontenc}    % use 8-bit T1 fonts
\usepackage{hyperref}       % hyperlinks
\usepackage{url}            % simple URL typesetting
\usepackage{booktabs}       % professional-quality tables
\usepackage{amsfonts}       % blackboard math symbols
\usepackage{nicefrac}       % compact symbols for 1/2, etc.
\usepackage{microtype}      % microtypography
\usepackage{xcolor}         % colors

\title{DeepSpeech models show Human-like Performance and Processing of Cochlear Implant Inputs}

\author{%
  Cynthia R. Steinhardt\thanks{corresponding author. Simons Society of Fellows Junior Fellow} \\
    Center for Theoretical Neuroscience \\
  Zuckerman Mind Brain Behavior Institute\\
  Columbia University\\
  New York, NY 10027 \\
  \texttt{cs4248@columbia.edu} \\
  \And
  Menoua Keshishian \\
  Department of Electrical Engineering \\
Zuckerman Mind Brain Behavior Institute \\
Columbia University \\
New York, NY 10027 \\
\texttt{mk4011@columbia.edu} \\
  \And
  Nima Mesgarani \\
  Department of Electrical Engineering \\
Zuckerman Mind Brain Behavior Institute \\
Columbia University \\
New York, NY 10027 \\
\texttt{nima@ee.columbia.edu} \\
  \And
  Kimberly Stachenfeld \\
  Google DeepMind \\
  Columbia University \\
  New York, NY\\
  \texttt{stachenfeld@deepmind.com} \\
}

\begin{document}

\maketitle

\begin{abstract}
  Cochlear implants(CIs) are arguably the most successful neural implant, having restored hearing to over one million people worldwide. While CI research has focused on modeling the cochlear activations in response to low-level acoustic features, we hypothesize that the success of these implants is due in large part to the role of the upstream network in extracting useful features from a degraded signal and learned statistics of language to resolve the signal. In this work, we use the deep neural network (DNN) DeepSpeech2, which processes audio inputs causally to perform phoneme prediction from spoken sentences, and use it as a paradigm to investigate how natural input and cochlear implant-based inputs are processed over time. We generate naturalistic and cochlear implant-like inputs from spoken sentences and test the similarity of model performance to human performance on analogous phoneme recognition tests. Our model reproduces error patterns in reaction time and phoneme confusion patterns under noise conditions in normal hearing and CI participant studies. We then use interpretability techniques to determine where and when confusions arise when processing naturalistic and  CI-like inputs. We find that dynamics over time in each layer are affected by context as well as input type. Dynamics of all phonemes diverge during confusion and comprehension within the same time window, which is temporally shifted backward in each layer of the network. There is a shift and reduction in amplitude of this signal during processing of CI inputs compared to natural inputs which resembles the timing and changes of EEG signals in the auditory stream. This reduction likely relates to the reduction of encoded phoneme identity and indicates similarity in representation of natural and CI inputs. These findings suggest that we have a viable model in which to explore the loss of speech-related information in time and that we can use it to find population-level encoding signals to target when optimizing cochlear implant inputs to improve encoding of essential speech-related information and improve perception.
\end{abstract}
\section{Introduction }
Deep Neural Networks have emerged as modeling framework capable of complex and human-like behaviors \citep{10.1016/b978-0-323-96104-2.00002-6} and recently have made particular strides in performing text-based and auditory language tasks \citep{10.48550/arxiv.1706.03762}. As they have gained human-like capabilities they have been increasingly compared to the representation and processing of the human  brain \citep{10.1016/j.neuron.2020.09.005, 10.48550/arxiv.2401.17671, 10.1371/journal.pbio.3002366}. While they omit certain biophysical details, these models are uniquely capable of capturing complex perception processes. Speech-to-text models (i.e. DeepSpeech2 \citep{amodei2015deep}) in particular have been applied to model naturalistic speech perception \citep{keshishian2021}. In this work, we investigate their applicability for modeling how speech perception is altered in the hearing-impaired patients with cochlear implants (CIs). 
CIs have restored hearing to over one million people around the world \citep{10.1371/journal.pone.0232421, 10.1121/10.0012825}. They encode sound with a limited number of electrodes (16 in the Advanced Bionic implant simulated in this study) which deliver pulses of current over time with amplitude modulated proportional to power in the spectra band being encoded per channel. This strategy enables an informative but limited audio channel compared to the full spectrum experienced by normal hearing subjects. Much of the efforts to address deficits of cochlear implants have focused on detailed biophysical modeling of voltage-driven activations of single-neurons in the ear itself. While this work has enabled CIs to better approximate the effect of sound on the lowest levels of auditory processing, these simulations are not able to model the entire hierarchy of auditory processing (from sound to phonemes to words to sentences), nor how auditory processing is altered over time and across regions of the brain \citep{10.1016/j.brs.2021.11.015}.

In this work, we aim to create a model system in which to investigate how electrical encoding of speech-related information at the cochlea affects speech comprehension at the word and phoneme level. Our specific contributions are as follows:

\begin{itemize}
    \item We develop a model of natural speech comprehension for patients with cochlear implants. Our approach is to combine (1) a “vocoder” model designed to mimic how a CI distorts the acoustics of an auditory signal and (2) a DeepSpeech2 model trained to convert speech to phonemes. The latter is a novel variant on the speech-to-text model, DeepSpeech2, which we dub Phoneme DeepSpeech2 (PhoDe). 
\item We validate this model, showing the model captures different aspects of error patterns in CI versus NH subjects in the types of errors made, the effect of background noise on error types, and phoneme confusion rates.
\item We find that the model shows similar characteristics of temporal processing on words and phonemes, in particular replicating delays in reaction time with CI inputs, background noise, consonant vs. vowel, and correct vs. error.
\item We replicate findings of an auditory hierarchy and find that dynamics across layers of the model recapitulate key features of neural dynamics across processing levels in the brain as measured with EEG in CI vs. NH subjects.
\end{itemize}

\begin{figure}[t!]
  \centering
  \includegraphics[width=0.9\textwidth]{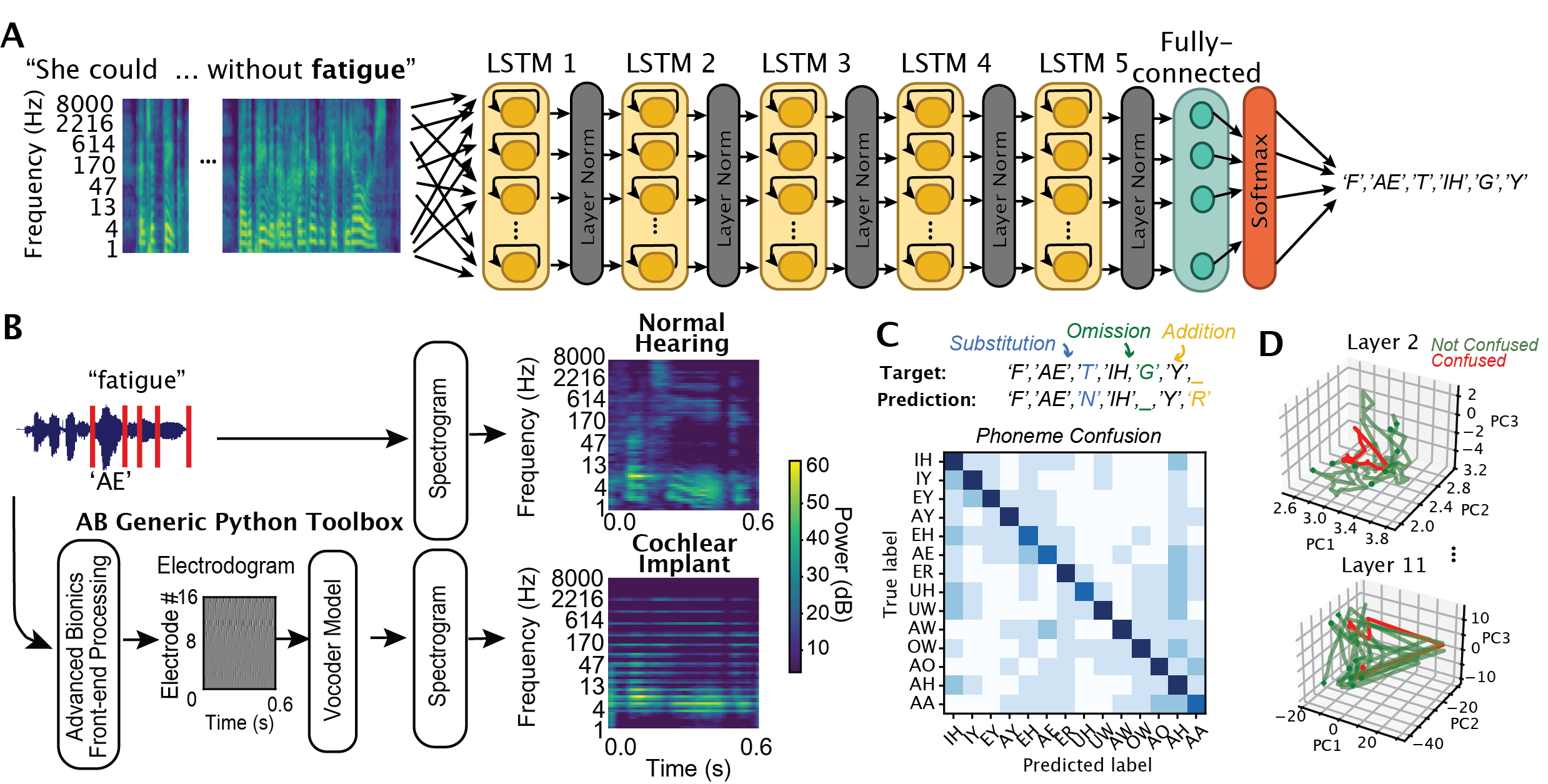}
  %\fbox{\rule[-.5cm]{0cm}{4cm} \rule[-.5cm]{4cm}{0cm}}
\caption{Auditory System Model and Input Generation. A. Phoneme DeepSpeech 2 (PhoDe) Network with 5 LSTM layers followed by a fully-connected layer was trained to process spectrograms of sentences from various speakers in LibriSpeech. B. Cochlear implant versions of inputs were made by running the audio through the front-end processing algorithm of an Advanced Bionics cochlear implant, then transforming the electrodogram via a biophysical model and filterbanks into a vocoded version of the speech. C. During testing, the output predicted sequence of phonemes was aligned to the (target) true phoneme utterances over time with a Levenshtein’s algorithm. The number of substitutions or confusions, omissions, and additions could be determined to find the phoneme confusion matrix. D. Example 3-D projection of activation during confusion (red) and non-confusion (green) of phonemes in network layers: Layer 2 (top) and Layer 11 (bottom). }
\end{figure}

\subsection{Related Work}
\subsubsection{Cochlear Implants}
Since the first use of electrical stimulation in a cochlear implant(CI) to restore hearing in 1957, CIs have  proliferated and restored hearing to over one million people around the world \citep{10.1371/journal.pone.0232421,10.1121/10.0012825} The success of CIs  inspired a migration of the invasive electrode hardware and pulsatile stimulation algorithm over to a  variety of devices \citep{10.1155/2018/1435030}, including retinal and vestibular implants for sensory restoration \citep{10.1586/erd.11.12,10.1016/j.jocn.2020.05.041,10.1172/jci.insight.128397}, spinal  cord stimulators for pain, and deep brain stimulators for treatment of motor and psychiatric disorders \citep{10.1186/s12938-020-00773-4,10.3390/jcm9103260}. While all these devices successfully aid in a range of restorative treatments, patient recovery remains  limited compared to normal function in each case \citep{10.1155/2018/1435030}\cite{10.1098/rstb.2013.0528}. Deficits are often attributed to significant  differences in the spatial targeting of neurons due to current spread from the electrodes \citep{10.3109/0954898x.2016.1171412}, unnatural  temporal synchrony of neurons due to pulse-locked activations \citep{10.1038/ncomms11238}, or the limited bandwidth of the signal  delivered, due to hardware limitations, especially in CI uses \citep{10.1121/1.1381538, 10.1044/jslhr.4005.1201}. With the largest patient population and over 60 years of use in real-world situations, the deficits of CIs in different types of noise, for speech tasks at the phonetic- \citep{10.1044/2018_jslhr-h-16-0463, 10.1044/2023_jslhr-23-00335}, word-, and sentence-level \citep{10.1097/aud.0000000000000253}\cite{10.1159/000315115} have been carefully characterized \citep{10.1371/journal.pone.0232421},  as has the psychophysics of the normal auditory system for equivalent tasks \citep{10.1121/1.1810292,10.1371/journal.pone.0079279,10.1016/j.joto.2020.12.001}. This combination of understanding the natural system and behavioral and recording data makes CIs an excellent test-bed for understanding differences in processing of information encoded with electrical  stimulation compared to natural inputs. 
\begin{figure}[t!]
  \centering
  \includegraphics[width=0.9\textwidth]{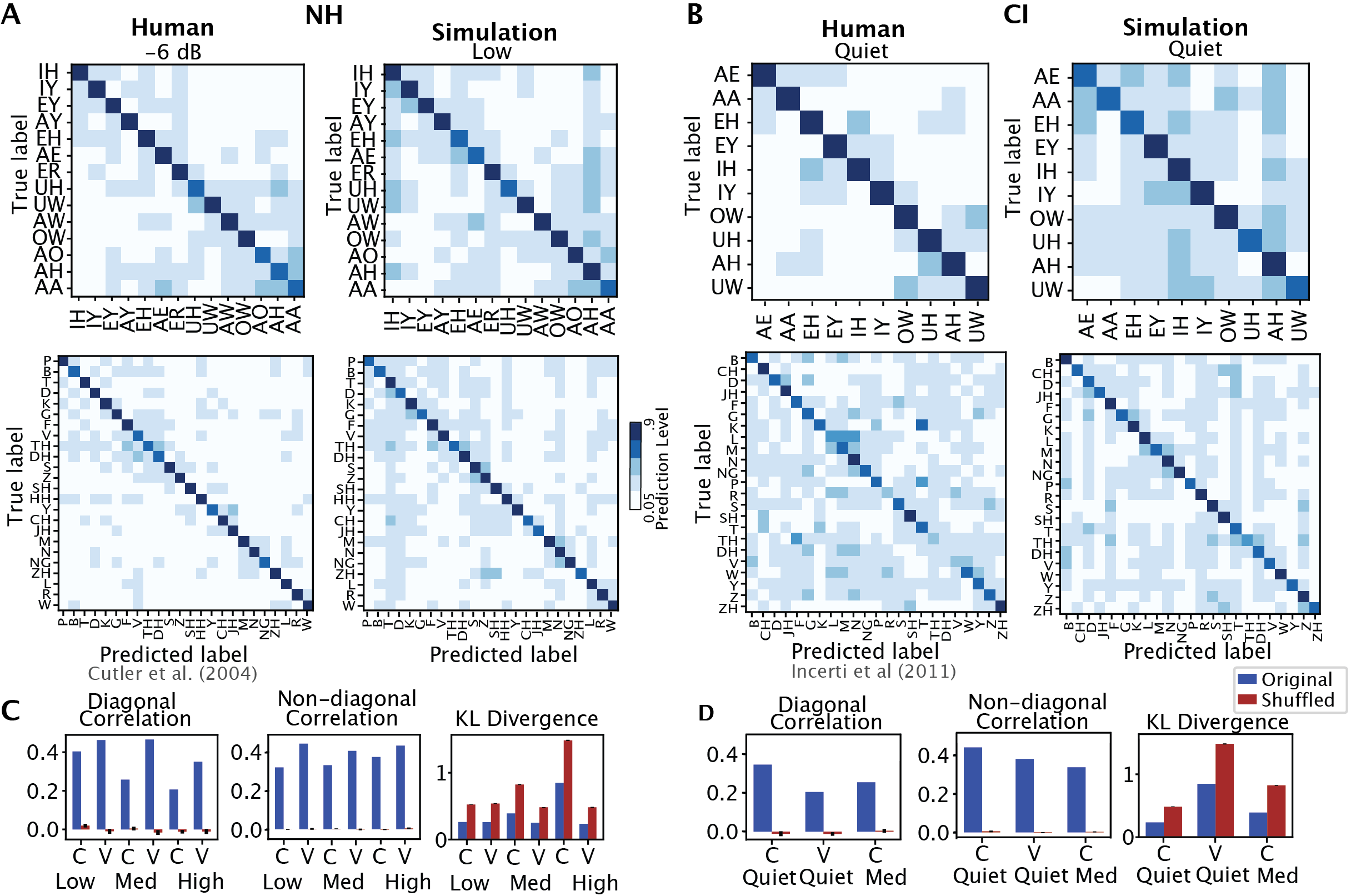}
  %\fbox{\rule[-.5cm]{0cm}{4cm} \rule[-.5cm]{4cm}{0cm}}
\caption{A. Human \cite{10.1121/1.1810292} NH consonant and vowel performance in -6 dB versus low noise simulation condition pattern of confusion. B.  Human\cite{10.1055/s-0031-1271950} and simulation comparison of CI listening in quiet for consonants and vowels. Shown at 5,40,70, and 90\% thresholding compared to normalized maximum prediction probability per phoneme. C. Diagonal correlation, non-diagonal correlation, and KL divergence between human and simulation confusion matrices for original matrices(blue) versus shuffling of the simulation matrix for 500 shuffles (red) at each noise level. D.Statistics for CI data. }
\end{figure}
\subsubsection{Local Biophysical Modeling of Electrical Stimulation}
Much of the efforts to address deficits of neural implants have focused on determining optimal stimulation parameters and hardware configurations for targeting desired local neuron populations using detailed biophysical modeling of voltage-driven activations of single-neurons in the complex three-dimensional geometries of the body \citep{10.3109/0954898x.2016.1171412, 10.1016/j.neurom.2023.04.471, 10.1111/ner.13037, 10.1007/s00221-011-2599-x}. While this body of work has improved the ability to tune parameters to optimize patient-specific outcomes, modeling neural responses over time locally and across brain regions \citep{10.1016/j.brs.2021.11.015} is intractable in these let alone the complex behaviors and deficits observed when using neural implants.

\subsubsection{ Deep Neural Networks as a Comparative Model System to Human Auditory Stream }
Auditory DNNs, specifically text-based large language models(LLMs) have been a particularly popular point of comparison since the success of transformer architectures at language task \citep{10.48550/arxiv.1706.03762}. Often, comparisons have addressed representational similarities between LLMs and the brain \citep{10.48550/arxiv.2401.17671, 10.1038/s42003-022-03036-1, 10.1038/s41562-022-01516-2, 10.1038/s41593-022-01026-4}. Each of these studies has revealed shared representational features between DNNs and the human auditory system, but they have also been limited particularly in addressing temporal processing similarities.  Some lacked temporal precision in comparisons due to use of functional MRI \citep{10.1038/s42003-022-03036-1, 10.1038/s41562-022-01516-2}.  Comparisons are often made on text-based LLM which differ significantly, especially in earlier processing stages from listening to spoken words \citep{10.48550/arxiv.2401.17671, 10.1038/s42003-022-03036-1, 10.1038/s41562-022-01516-2, 10.1038/s41593-022-01026-4}. Additionally, models process inputs in biologically implausible manners that lack causality \citep{10.1038/s41593-023-01468-4} in how inputs are integrated to perform speech comprehension. As a result, certain model architectures share more representational similarity to the human auditory stream \citep{10.1371/journal.pbio.3002366}. We choose to focus our efforts on a DeepSpeech2 model that has been shown to share the temporal processing hierarchy of phonetic and semantic content \citep{keshishian2021} and causality with the human auditory system \citep{10.1523/jneurosci.3684-10.2011, 10.1101/761593}. 

In this work, we aim to create a model system in which we can investigate how electrical encoding of speech-related information in the cochlea affects speech comprehension at the word and phoneme-level. 
Previous recording \citep{10.1523/jneurosci.1699-16.2016} and simulation studies \citep{10.1088/1741-2552/ad36e2} indicate pulsatile stimulation produces different encoding patterns than natural inputs in higher-order auditory cortex \citep{10.1109/embc48229.2022.9871812}. However, we hypothesize non-identical inputs could also produce similar deep layer responses \citep{10.1038/s41593-023-01442-0} and therefore better speech encoding without producing identical cochlear activity to the healthy cochlea which is intractable with current technologies \citep{10.1101/2021.08.18.456731}.

\begin{figure}[t!]%[ht]
  \centering
  \includegraphics[width=0.9\textwidth]{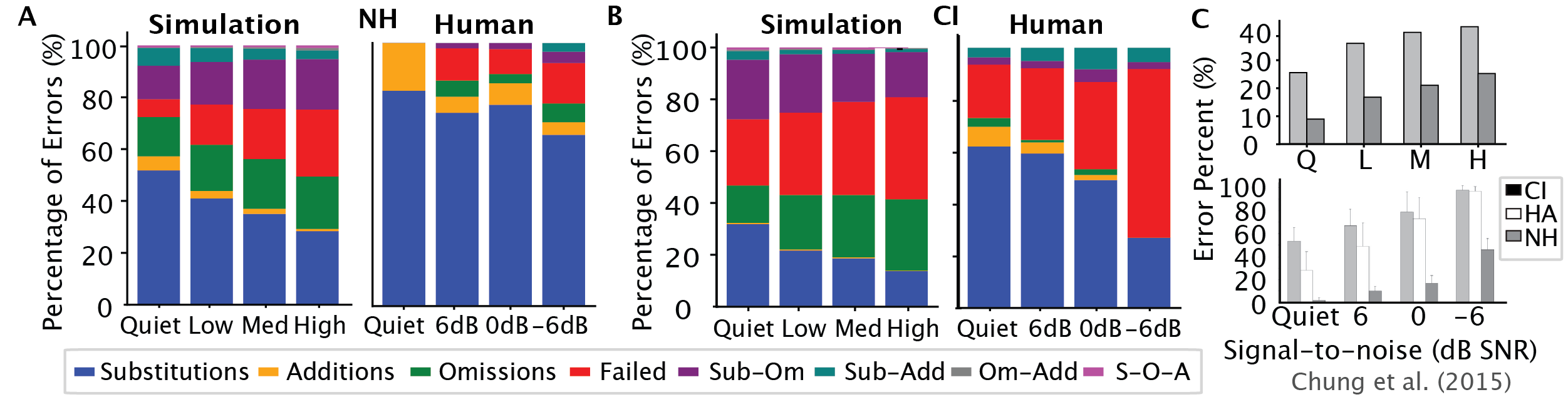}
  %\fbox{\rule[-.5cm]{0cm}{4cm} \rule[-.5cm]{4cm}{0cm}}
  \caption{ Error Rate Comparison. Errors types were Substitution(blue), Addition(yellow),Omission(green),Failed(red), Sub-Om (purple), Sub-Add(teal),Om-Add(grey),S-O-A(all three,pink). Percent of each error made per word by simulation(left) versus humans (right) in A. NH condition and B. the CI condition. C. The percent of correctly identified phonemes at all noise levels by (top) the network and (bottom) human subjects. All comparisons were made to data from \citep{10.7874/jao.2015.19.3.144}. } 
\end{figure} %\cite{10.7874/jao.2015.19.3.144} 

\begin{figure}[ht]
  \centering
  \includegraphics[width=0.9\textwidth]{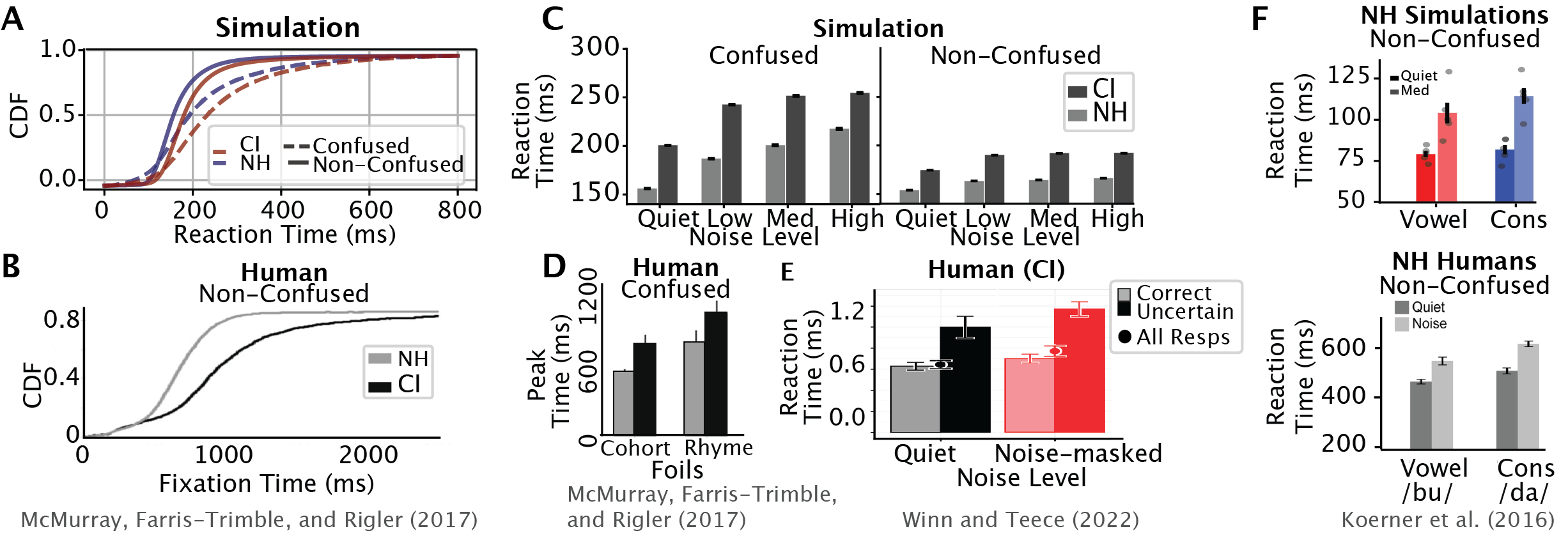}
  %\fbox{\rule[-.5cm]{0cm}{4cm} \rule[-.5cm]{4cm}{0cm}}
  \caption{Reaction Time Comparison. A. CDF of reaction times for all phonemes for CI(red), NH(blue),  confused (dashed), non-confused(solid). B.D. from \cite{10.1016/j.cognition.2017.08.013}. B Time to fixate to image of heard word for NH than CI. C. Reaction time for confused (left) versus non-confused(right) phonemes in CI(black) and NH(grey) conditions for the simulations with increasing noise level. D. Time to fixate image for foils -cohort  (words with a similar starting phoneme e.g. wizard/whistle)  and rhyme for NH and CI subjects. E. CI subject reaction time for certain and uncertain word predictions in quiet(black) and noise(red) from \cite{winn2022effortful}. F. Reaction time of model for vowels(red) versus consonants(blue) for non-confused phonemes in quiet(dark) and medium noise(light) G. Reaction time for NH humans for vowel or consonant identification in quiet and noise (4-talker babble) from \cite{10.1016/j.heares.2016.06.001}.  } 
\end{figure}

\begin{figure}[ht]
  \centering
  \includegraphics[width=0.9\textwidth]{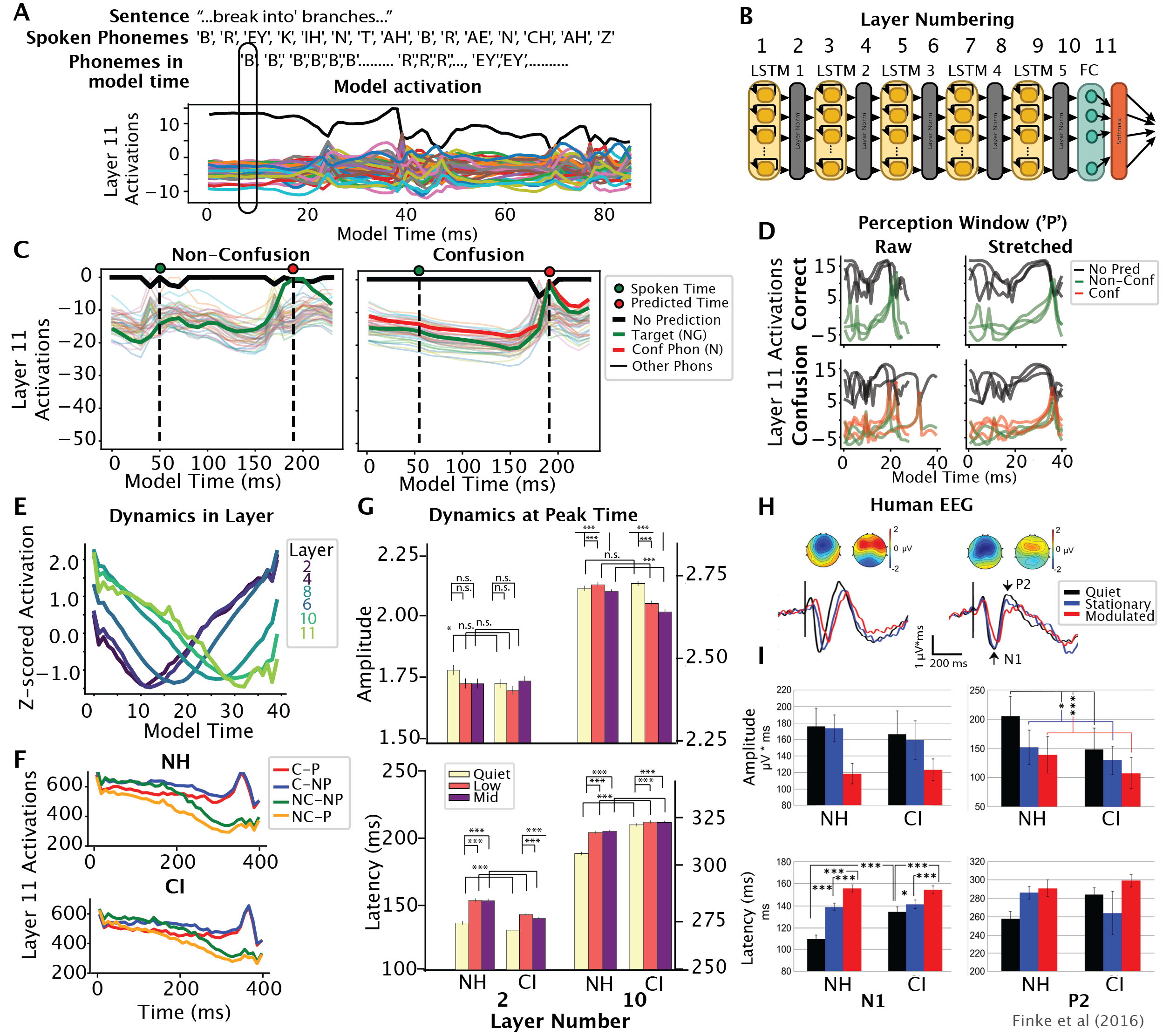}
  %\fbox{\rule[-.5cm]{0cm}{4cm} \rule[-.5cm]{4cm}{0cm}}
  \caption{Differences in Dynamics during Confusion and Non-confusion. A. Model activity can be parsed into time from phoneme onset to phoneme prediction per phoneme. B. Numbered layers in the network as referenced in C-G. C. Layer 11 activations for non-confusion of ‘NG’(green) (left) and confusion with ‘N’ (red). Other phoneme-related activations shown in various colors with thinner lines. The no prediction signal(black) dips during phoneme onset (green circle) and phoneme prediction(red circle). D. Raw activation in Layer 11 (left) versus utterance windows interpolated to the same length (40 model time points/400 ms). E.  Z-scored distance in PC space between dynamics when phonemes are NC colored purple to light green by depth of layer in the model for NH inputs. F. Distance in PC space between dynamics during processing of NH (top) and CI (bottom) inputs during utterances that were C-P(red),C-NP(blue), NC-NP(green), and NC-P(yellow). G. Change in amplitude and latency of the peak response time with increase in noise (quiet-sand, low-peach,medium-purple) for Layer 2 and Layer 10 which have different average latencies of response. H. ERPs from whole-brain human EEG to words in quiet, stationary noise, or modulated noise in \cite{10.1159/000452123}. N1 and P2 times were found at about 130 and 250 ms delays. I. Amplitude and latency of ERP peak response under quiet and noise conditions for NH and CI listeners.}
\end{figure}
\section{Methods}
\subsection{Phoneme DeepSpeech2 Model (PhoDe)}
We used a DeepSpeech2 architecture trained end-to-end to convert spectrograms to English phonemes using CTC Loss \citep{10.1109/icassp.2015.7178964}(Figure 1A).   Our model consisted of five causal LSTM layers, a fully-connected layer, and a softmax layer with batch normalization following each LSTM layer(Figure 1A, Supp. Table 1). 

\subsection{Constructing Cochlear Implant-like Inputs} 
Vocoders have been developed as a research tool for imitating the distortion of audio experienced by CI users due to the limitations of CI hardware, such as limited electrode count, frequency shift, and distortions due to current spread \citep{10.1177/2331216519866029}. For the CI version of each input, audio was processed using the Advanced Bionic Generic Toolbox front-end processing to produce the stimulation per electrode channel (electrodogram)\cite{ABGenericPythonToolbox}. Then, a biophysical model followed by a filterbank-based vocoder was used to reconstruct an equivalent  CI audio, and the same spectrogram procedure was used to make an equivalent CI version of each test audio sequence in the 10,028 sentence set(Figure 1B). 

\subsection{Dataset \& Training}
Training and evaluation data consisted of 64-channel spectrograms created from natural speech from the LibriSpeech \citep{10.1109/icassp.2015.7178964} dataset. LibriSpeech includes 1000 hours of speech read by male and female speakers. The model was trained with supervision to predict phonemes from 280,000 sentences. The DeepSpeech2 model was trained only on normal hearing spectrograms; thus, recapitulating behavior on vocoded CI spectrograms requires a degree of out-of-distribution generalization. Speech recordings were either in quiet (unaugmented) or augmented using the Sound eXchange (SoX) backend from torchaudio with background noise, reverberations, frequency masking and stretching and pitch shifting inputs. To create different noise levels (low, medium, and high), we increase the range of parameters of each of these augmentations (Supp. Table 2). To improve model generalization, the model was trained with quiet and low-level augmentations. 

\subsection{Alignment of Model Predictions and Utterance Window Isolation}
The alignment of spoken and predicted phonemes was important for two reasons. Patterns of error confusion were used as a metric of comparison to human data, and, after alignment, the window between phoneme speech onset and prediction time of the phoneme or the confused phoneme was used to analyze differences in dynamics during neural processing of the phoneme. We use the spoken phoneme order without spaces as a target sequence and the output of our model without spaces or blanks as the predicted sequence. The Levenshtein algorithm \citep{levenshtein1966} was used to add insertions to both sequences to best align them based on the chosen phoneme. 
The utterance window, the time between phoneme onset from the audio segmentation and prediction of a phoneme was recursively isolated for each phoneme to use for analysis of dynamics for phonemes not paired with an insertion.  Due to a combination of hallucinations, omissions, and confusions, alignment by the Levenshtein algorithm alone did not recapitulate the spoken and predicted pairings. Thus, a secondary correction algorithm was used to ensure the spoken phoneme from the alignment precedes the predicted phoneme, which moved insertion locations if this was not the case. In an increasing number of sentences, as noise was introduced, predictions began to proceed target times. Sentences in which this occurred for the NH or CI condition were excluded from further analyses. 

\subsection{ Latency and Amplitude of Dynamics }
To assess for dynamical signatures of preserved encoding, we analyze dynamics per layer of the model in the time between phoneme onset and prediction, as this is the window in which information must be encoded and interpreted by the network.
Evoked response potentials(ERPs), changes in electroencephalogram (EEG) after a stimulus onset have been characterized for a variety of sensory processing tasks and cognitive disorders; changes in amplitude and latency of response have been linked to differences in sensory processing \citep{10.4103/0972-6748.57865}. Changes in ERPs are thought to reflect synchronous changes in postsynaptic potentials that occur within a large population of local pyramidal neurons \citep{10.1093/acprof:oso/9780195050387.001.0001} and therefore reflect local processing. To get equivalent ERPs for comparisons, studies are typically designed with identical duration stimuli. Here, we time-lock to phoneme onset and interpolate the response to each phoneme to get an equivalent window. Distance between dynamics in PC space over time is used as a measure of the local change in activity. The time when the distance is the smallest is considered the time of peak responses. Then, latency and amplitude of the maximum change in response compared to the first 3 time points of the utterance window (baseline) is calculated in each later. These measures are compared to human EEG response data \citep{10.1159/000452123}.

\section{Results}
We conduct experiments on PhoDe to determine whether its performance on naturalistic and CI-like inputs is similar to normal hearing(NH) subjects and CI users performing analogous experiments. We find that our model can recapitulate a similar proportion of errors, changes in reaction time, and confusion patterns with NH versus CI inputs and as the noise level of the input increases. Moreover, we inspect this model to unpack temporal processing per network layer. We find a time-locked response in each model layer that shares similarities to human ERPs. This input changes with context before the phoneme, confusion of phoneme, and the input type and noise level. We find increases in latency and reductions in amplitude of responses in the network activity, like those observed in human EEG, when subjects respond to CI inputs with increasing noise levels compared to naturalistic inputs in quiet. We hypothesize these network signals indicate the correct encoding of phoneme identity and conclude that we can use them to optimize cochlear implant encoding strategies for future improvement of stimulation algorithms.

\subsection{Comparison to Human Speech Perception}
A unique feature of PhoDe is that it not only is capable of processing speech inputs in the form of spectrograms like prior DNNs\cite{10.1038/s41593-023-01442-0} but also that due to its more biophysically realistic architecture, it is constrained to processes inputs like human listeners, causally overtime. Thus, the model is expected to show contextual effects processing effects similar to humans. Training PhoDe to predict phonemes allows us to compare errors to human studies at the phoneme-level, often considered the smallest unit of speech, up to the word-, sentence-, and semantic-level. Here, we compare model performance to three aspects of human performance: (1) the pattern of phonemes confusions, (2) the frequency and types of errors when processing words, and (3) reaction time.

\subsubsection{Similarity in Phoneme Confusion Pattern with Noise and under CI Conditions}

Given PhoDe predicts speech at the phoneme-level, the most direct comparison that can be made is phoneme confusion across all sentences. The target, the sequence of phonemes spoken, was aligned to the predicted phonemes over time, as described in Methods 2.4. Then, phoneme confusion across the same test sentences could be measured in the NH and CI conditions with increasing noise levels.  We compared our results for NH conditions against Cutler et al. (2004) \citep{10.1121/1.1810292}. For the CI condition, we compare vowel confusion to Munson and Nelson(2003)\cite{munson2003} and consonant confusion to Incerti et al. (2011)\citep{10.1055/s-0031-1271950} (Figure 2, Supplemental Figure 1).  Across NH human studies, there is a baseline level of variability due to differences in tests and individual variability; differences only increase in CI test conditions where there is an added variability due to differences in experience with the implant, CI coding strategies, and hearing impairment \cite{valimaa2002phoneme}. Thus, we aim to determine whether prominent confusions are shared with prominent ones in population confusion matrices, although those may also differ across studies \cite{rodvik2018consonant}\cite{valimaa2002phoneme}.

We find some of the main confusions are present and the effect of a CI input and noise resembles human data. less confusion of ‘S’,’R’,’K’, and ‘T’ occurs, and ‘K’/’T’,‘L’/’M’, ‘M’/’N’, and ‘P’,’T’ confusion patterns are common, like human data\cite{rodvik2018consonant}\cite{valimaa2002phoneme}. ‘AH’ and ‘UH’ confusions are also common in our model like in human data\cite{rodvik2018consonant}. We see a bias towards the prediction of ‘T’,’S’, and ‘AH’ that is not apparent in human data. This may derive from learned statistics of language. Like human performance, added noise does not change prominent confusion patterns; it only amplifies existing confusions \cite{10.1121/1.1810292} (Supp. Fig. 1). With CI inputs, like human data, the most amplified confusions are ‘TH’/’DH’/’SH’ and /’L’/’M’/’N’/’NG’ for consonants and ‘AA’/’AH’/’AO’ for vowels which are also present in human data \cite{10.1055/s-0031-1271950}; ‘AH’ and ‘UH’ confusions also become more prominent in the CI case like human data (Figure 2A-B, Supp. Fig. 1). 

We find statistically significant similarity between human and simulated confusion patterns across NH and CI simulations and with increasing noise levels(Figure 2C-D, Supp. Fig. 3). Correlations between 0.2 to 0.45 of the diagonals of the normalized confusion matrices indicate similar relative confusion of phonemes (shuffle statistic, p <0.001, Figure 2C-D). KL divergence of the off-diagonals per phoneme and correlation of the off-diagonals were also statistically significant, indicating similar pattern of confusion with other phonemes per phoneme in NH and CI conditions (shuffle statistic, p <0.001, Figure 2C-D). 

\subsubsection{Capturing Error Patterns During  Phoneme Recognition in Words }
Because the model is trained to produce sequences of phonemes, we can evaluate errors at the sentence- and word-level. We compare the pattern of errors to those from the Chun et al.(2015) \cite{10.7874/jao.2015.19.3.144} experiment, which measured the number of phoneme errors per individual tri-phoneme monosyllables in increasing levels of noise in NH and CI conditions. (Figure 2A). 

We find similar patterns of errors by PhoDe for tri-phoneme words in test sentences. The main type of error was substitution(blue) for NH, like in the human data; as noise increased, the relative amount of omission (green), failed(red) and sub-om(purple) errors increased, while the percent of additions(yellow) and substitutions(blue) decreased; a relatively smaller number of Sub-Adds(teal) also increased with noise (Figure 3A). 
For CI inputs, our model showed a relatively lower portion of substitutions than NH inputs, like the data, and the number of failed errors increased with noise, while the number of sub-om, and additions decreased with noise. Additions also made up a relatively smaller percentage of the errors overall for CI inputs (Figure 3B).Overall, the percentage of errors increases with noise for both CI and NH inputs, and errors in the CI case were substantially higher than NH errors across all noise levels (Figure 3C).

Potentially, because our model processes words in continuous speech instead of individual words, we see a higher level of mixed errors even in quiet than in the human study.  Our model overall also showed more omissions than human data, which may contribute to how low substitution errors are relative to other errors.  We note that the percentage of NH and CI errors in the model was consistent with Finke et al.(2016)\cite{10.1159/000452123} where word error per sentence in continuous speech was measured (Supp. Fig. 2A).

\subsubsection{Capturing Change in Reaction Time with Noise and under Cochlear Implant Conditions }
One unique benefit of PhoDe is how it is constrained to process inputs causally, like humans. So, we hypothesize that behavioral metrics, such as reaction time, which have been shown to increase with confusion and difficulty of task in humans may also increase in our model.  We operationalize reaction time(RT) as the time from phoneme onset to prediction of a phoneme for each phoneme in the test set for our model.  We can then measure RT in the CI(red) and NH(blue) condition and during confusion(dashed) versus non-confusion(solid)(Figure 4A, Supp. Fig. 2B). 

The CDF of RT is compared to human time to fixation of a visual image of a spoken word from McMurray et al. (2017)\cite{10.1016/j.cognition.2017.08.013} (Figure 4B). Like human data, we see faster RT to NH inputs than CI inputs. During confusion(C) and non-confusion(NC), RT to NH inputs is faster than CI inputs, and RT to NC inputs is faster than C inputs(Figure 4C).  As noise increases, RT increases for NH and CI conditions (Figure 4A,C). This emulates data from McMurray at al. that shows for foils (confused target) RT is faster for NH than CI (Figure 4C, left). In Winn and Teece (2002)\cite{winn2022effortful}, CI users also performed word recognition in sentences in quiet and noise, and RT significantly increased with noise and was slower when subjects were uncertain or confused in both conditions. This is also consistent with model performance.  Note, in a similar study, Finke et al.(2016), CI RTs were significantly higher but the increase in RT with noise was not significant(Supp. Fig. 2), so the strength of the noise effect varies with the task. Finally, we determine if vowel and consonant RT differences are present in the model. Human studies show RT to consonants is more affected by noise, and RTs for consonants are higher than for vowels \cite{10.1016/j.heares.2016.06.001} (Figure 4F bottom). We see both of these effects when looking at the five vowels and consonants with the fastest RTs(Figure 4F top). In the full phoneme set, consonant RT were still more affected by noise than vowel RTs, but some baseline vowel RTs were higher than consonant RTs(Supp. Fig. 2 C,E). Overall, these results show PhoDe shares phoneme-specific RT effects with humans under NH and CI conditions (faster RT in NC, faster RT for NH inputs, and relatively stronger effects of noise on consonant RTs), indicating many similarities in processing of phonemes over time.

\subsection{Identifying ERP-like Dynamical Signatures of Phoneme Recognition} 
PhoDe shares several performance effects with human subjects and is processing inputs that share many features of natural and CI inputs to the auditory system. Thus, we assess whether there is activity within the model that indicates successful encoding of speech information with artificial inputs.  To do so, we isolate the time between phoneme speech onset, as determined by the speech aligner, and the time of phoneme prediction, which we call the utterance window during confusion(C) and non-confusion/comprehension (NC) of phonemes (Figure 5A-C). All utterance windows were interpolated to a fixed length and z-scored to account for differences in layer size(Figure 5D). Then, we could assess for delays and amplitude changes in fixed response windows. Dynamics in each layer become most similar, as measured by distance in PC space, at distinct times after phoneme onset. The time at which dynamics became most similar was $t_{peak}$. $t_{peak}$ shifts back in time with the depth in the network (Figure 5E). Additionally, we find that $t_{peak}$ is modulated by whether the input was C or NC and contextually whether the phoneme was probable(P) or not probable(NP). P phonemes (yellow and red) converging in dynamics sooner and NC-P dynamics reaching minimal distance in all layers across utterances (Figure 5F, Supp. Fig. 3B). The latency of $t_{peak}$ and amplitude of deviance from baseline activity ($t = 0-30$ ms) are both modulated by increases in noise and whether the inputs in NH and CI (Figure 5G).

In human EEG studies, the ERP to words in continuous speech were recorded with increasing levels of noise (Figure 5H). The ERP showed differences in modulation of latency and amplitude of the N1(00) and P2(00) signal for NH and CI subject and with increasing noise.  The changes closely resemble changes in activity at peak time in Layers 2 and 4 and Layers 8 and 10 respectively (Figure 5G, Supp. Fig. 3C). Both the N1 and P2 showed an increase in latency with noise and reduced amplitude; additionally latencies were longer for CI users across conditions and amplitudes were stronger. Differences in amplitude were more significant in P2 than in N1, and differences were more significant for N1(Figure 5H). These results are compatible with changes in PhoDe dynamics in Layers 2/4 and 8/10. The main difference we find is that differences in latencies are significant across conditions for our model(Supp. Fig. 3C). However, the magnitude of differences in latencies reduced in Layers 8/10, further reflecting the timing and modulation of human EEG activity during auditory processing. 
These findings reveal a signature of confusion and non-confusion with similar timing for all phonemes. It shares similarities with human ERP N1s and P2s during NH and CI conditions. This may be usable as a marker of comprehension of spoken phonemes and therefore as an optimization target. We also find that differences arise in Layer 2 and propagate in the network(Supp. Fig. 3A and surrounding discussion) and find specific time windows for potential intervention in each layer that could be used for future work in optimizing inputs to maximize encoding of information.

\section{Discussion} 
Our findings support PhoDe as a potential clinical model for investigating how auditory information is processed over time in normal hearing and cochlear implant conditions. Having the ability to model electrical stimulation-based inputs throughout the auditory processing hierarchy over time in models performing complex tasks, such as speech recognition, may allow us to find neural signatures of speech comprehension that may be used to improve cochlear implant algorithm performance. Additionally,  because of the similar encoding approach of cochlear implants and other devices, the understanding gained about network-level processing of these artificial inputs applies to other neural implants, such as retinal implants or deep brain stimulators. 

We have several limitations in our ability to model and compare to human experiments. There is variability in performance of NH subjects and  CI users, especially in phoneme confusion, which has been attributed to differences in implant location, remaining inner ear health, age, and cognitive factors \cite{10.1371/journal.pone.0232421}. We also did not have a precise implant placement to replicate and pulled data from multiple studies of speech perception and the word and phoneme level that uses different types of noise. Thus, we cannot directly compare model performance. Additionally, our vocoder model is also not an accurate representation of CI inputs. In future work, we could replace the inputs with a more biophysical model of neural activations based on human CI placement maps. This paper aims to propose a model system to investigate differences in human processing of electrical and CI inputs and find signatures to use for optimizing population-level encoding with electrical inputs. We feel these limitations do not significantly affect our findings, although comparisons would likely improve with further biophysical accuracy.

\begin{ack}
This work was supported by a grant from the Simons Foundation ( 965377 CRS). We thank Andrea  Weber for sharing data from Cutler et al. (2004).

% More information about this disclosure can be found at: \url{https://neurips.cc/Conferences/2024/PaperInformation/FundingDisclosure}.
\end{ack}

% Do {\bf not} include this section in the anonymized submission, only in the final paper. You can use the \texttt{ack} environment provided in the style file to automatically hide this section in the anonymized submission.
% \end{ack}

% \section*{References}

% References follow the acknowledgments in the camera-ready paper. Use unnumbered first-level heading for
% the references. Any choice of citation style is acceptable as long as you are
% consistent. It is permissible to reduce the font size to \verb+small+ (9 point)
% when listing the references.
% Note that the Reference section does not count towards the page limit.
% \medskip

% {

% Continue for all references, making sure to format each entry with the correct details.

% \bibliographystyle{plain}
% \small
% \bibliography{neurips_2024}

% [1] Alexander, J.A.\ \& Mozer, M.C.\ (1995) Template-based algorithms for
% connectionist rule extraction. In G.\ Tesauro, D.S.\ Touretzky and T.K.\ Leen
% (eds.), {\it Advances in Neural Information Processing Systems 7},
% pp.\ 609--616. Cambridge, MA: MIT Press.

% [2] Bower, J.M.\ \& Beeman, D.\ (1995) {\it The Book of GENESIS: Exploring
%   Realistic Neural Models with the GEneral NEural SImulation System.}  New York:
% TELOS/Springer--Verlag.

% [3] Hasselmo, M.E., Schnell, E.\ \& Barkai, E.\ (1995) Dynamics of learning and
% recall at excitatory recurrent synapses and cholinergic modulation in rat
% hippocampal region CA3. {\it Journal of Neuroscience} {\bf 15}(7):5249-5262.
% }

%%%%%%%%%%%%%%%%%%%%%%%%%%%%%%%%%%%%%%%%%%%%%%%%%%%%%%%%%%%%
\newpage
\appendix

\section{Appendix}

\subsection{Additional Model Details}
All models were implemented in PyTorch on the training set of the LibriSpeech corpus \cite{10.1109/icassp.2015.7178964}. All model layers contained 500 neurons, except for the last layer which contained 41 units related to the 39 English phonemes, the blank, and space token predictions. We used the Adam optimizer (learning rate: 1.5e-4, weight decay: 1e-5) and a batch size of 64. Training was of all models were performed on NVIDIA A40 and L40 GPUs (one per  training) at the internal cluster of our organization. Each epochs of training took approximately 2.5 hours for each model, totaling 30 epochs with  8787 steps per epoch.  

\subsection{ Evaluation of Prediction Performance}
\subsubsection{ Error Metrics}
To compare to the Chun et al(2015) \citep{10.7874/jao.2015.19.3.144} experiment, phoneme errors present after final alignment were re-attributed to the word they were uttered in using the spoken word and phoneme segmentation information (Figure 2). Error rate was counted at the word-level, like in the study, where if only a substitution occurred in a word, this was considered one substitution error in the total count, but if a substitution and omission occurred, this was considered only substitution-omission error. The experiment used tri-phoneme words, so we restrict this measure only to words containing three  phonemes (Figure 2A-B). Categories are self-explanatory, except for failed, meaning three errors on these three phoneme words.
\subsubsection{Reaction Time} 
Using the utterance windows, we operationalize reaction time for each phoneme as the time from phoneme onset to the prediction time of the output in millisecond. For vowel and consonant reaction times the average length of a vowel or consonant utterance was subtracted from this difference as the reaction time to prediction \cite{ladefoged2006course}. 
\subsubsection{  Human Phoneme Confusion Data Comparisons} 
In order to quantify the extent to which our model captures the same pattern of errors as humans, comparisons were made to four different studies of English phoneme perception. The confusion matrices from Cutler et al. (2004) \citep{10.1121/1.1810292} for normal hearing subjects listening to consonants and vowels at three noise levels (+6 dB, 0 dB, -6 dB) are used to compare to simulations at low, medium, and  high noise for natural inputs (Supp. Table 2). Two cochlear implant studies of confusion were used. We use vowel  confusion data from Munson and Nelson (2003) \citep{munson2003} of speech in quiet for better listeners as a comparison  to the simulation in quiet conditions. We also use consonant confusion from Incerti et al. (2011) \citep{10.1055/s-0031-1271950} of  speech in quiet and with 8-talker babble as a comparison to our model processing CI inputs at quiet and  medium noise levels.

\subsubsection{Confusion Matrix Similar Metrics}
For all phonemes that did not have an insertion in the final alignment the pattern of confusion could be measured. For comparison of simulation confusion to human confusion in each study, only the phonemes present in that study were included in the matrix. We use several similarity metrics. The correlation between the diagonal elements of the matrices only were used as a measure of the similarity in relative confusability of each phoneme. The correction of the off-diagonal only and the KL-Divergence per row were used as a measure of similarity in pattern of confusion per phoneme. Overall correlation and Manhattan distance were also used to measure overall performance similarities. All comparisons were made on the row-normalized matrices. A shuffle comparison was made with 1028 shuffles of paired rows to evaluate significance.
\subsection{Additional Model Interpretability Methods}
\subsubsection{Decoder of Intact Encoding of Spoken Phonemes over Time}
To assess the level of preserved information about the original audio input per layer, we use a linear decoder(SVD) to decode the spoken phoneme at each time point from the activation outputs of the model. We consider the phoneme as being present at each time point between the phoneme onset and offset based on segmentation. Decoding was performed per layer for 100 concatenated sentences with 10 80-20 cross-validated splits of the data.
\subsection{Determine Context Effects from Previous Phonemes}
A bi-gram model was made by using the full LibriSpeech dataset to directly count statistic of occurrence. A phoneme was considered probable (P) if it was in the top 10 \% of phonemes followed by the proceeding phoneme and not probable (NP) if it was in the remaining 90 \%.
\subsection{Constructing Comparable Dynamics Windows}
The utterance of each phoneme varies speaker-to-speaker and sentence-by-sentence. To make utterances comparable, the time of prediction of the model was found based on the time of peak of the predicted phoneme-associated neuron activations in Layer 11, which predicts a phoneme by producing a sharp increase in activity of one of the output-associated neurons. Then, the utterance time series was interpolated to a fixed length from two time points between the phoneme onset time to 0.2*(utterance length) time points to 0.25*(utterance length) time points after prediction time to produce a fixed length of 40 model time steps or 400 ms in real-time (Figure 5D). Phoneme utterances were categorized as confused(C) or non-confused(NC) and probable(P) or not probable(NP). Up to 50 exemplars of each category (C-P,C-NP,NC-P,NC-NP) were collected for each phoneme in the test set. Phoneme comparisons were not made if there were not at least 2 exemplars in each category, leaving 34 phoneme comparisons.

\subsection{PC Space Visualization Details}
To visualize and measure differences in dynamics in the network a principal component analysis (PCA) was used to project activations over time into a shared space. For dynamics comparisons, distance was measured in the full PC space dimensionality. Distance metrics were compared in the projection into the shared PC space of all NH and CI input responses that were C or NC in the test set. For visualizations of the difference in dynamics in NH and CI conditions in Figure 5F and the Supplemental Figures 3 and 4, data were projected into a 3-D space of the PC space for NH responses.

All code used for training the model and these analyses is available at https://github.com/ANONYMOUS %CSteinhardt153/CIDNNProject

All figures with errorbars show the mean and S.E.M of the data.

\begin{figure}[h]
  \centering
  \includegraphics[width=1\textwidth]{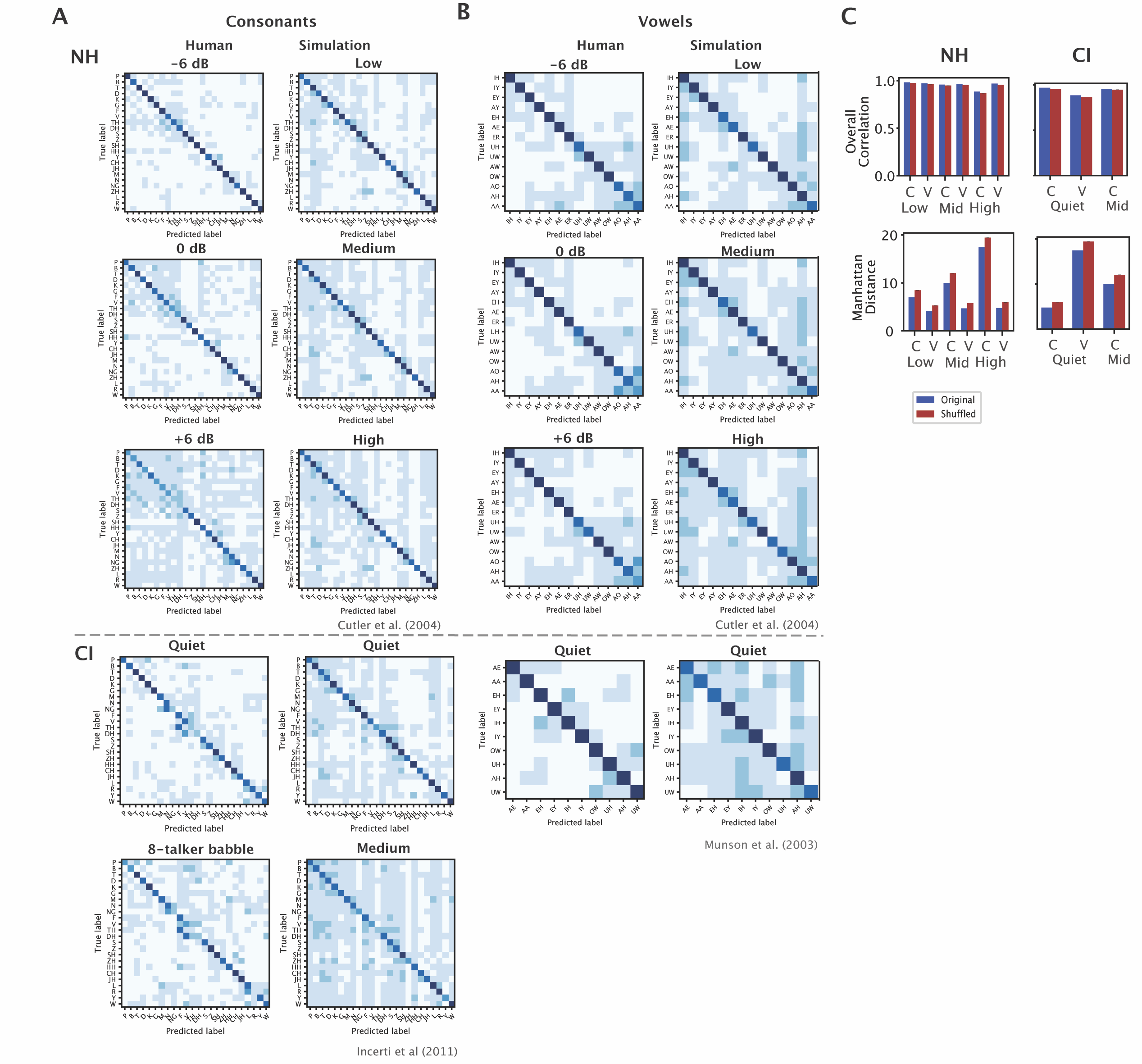}
  %\fbox{\rule[-.5cm]{0cm}{4cm} \rule[-.5cm]{4cm}{0cm}}
  \caption*{Supplemental Figure 1. Comparison of human(left) and simulated(right) NH (top) and CI(bottom)  confusion matrices for A. consonants and B. vowels. during increasing noise levels. C. Overall matrix  correction and Manhattan distance between human and simulation confusion matrices for original matrices(blue) in A versus paired shuffling of the simulation matrix for 500 shuffles (red).}
\end{figure}

\begin{figure}[h]
  \centering
  \includegraphics[width=1\textwidth]{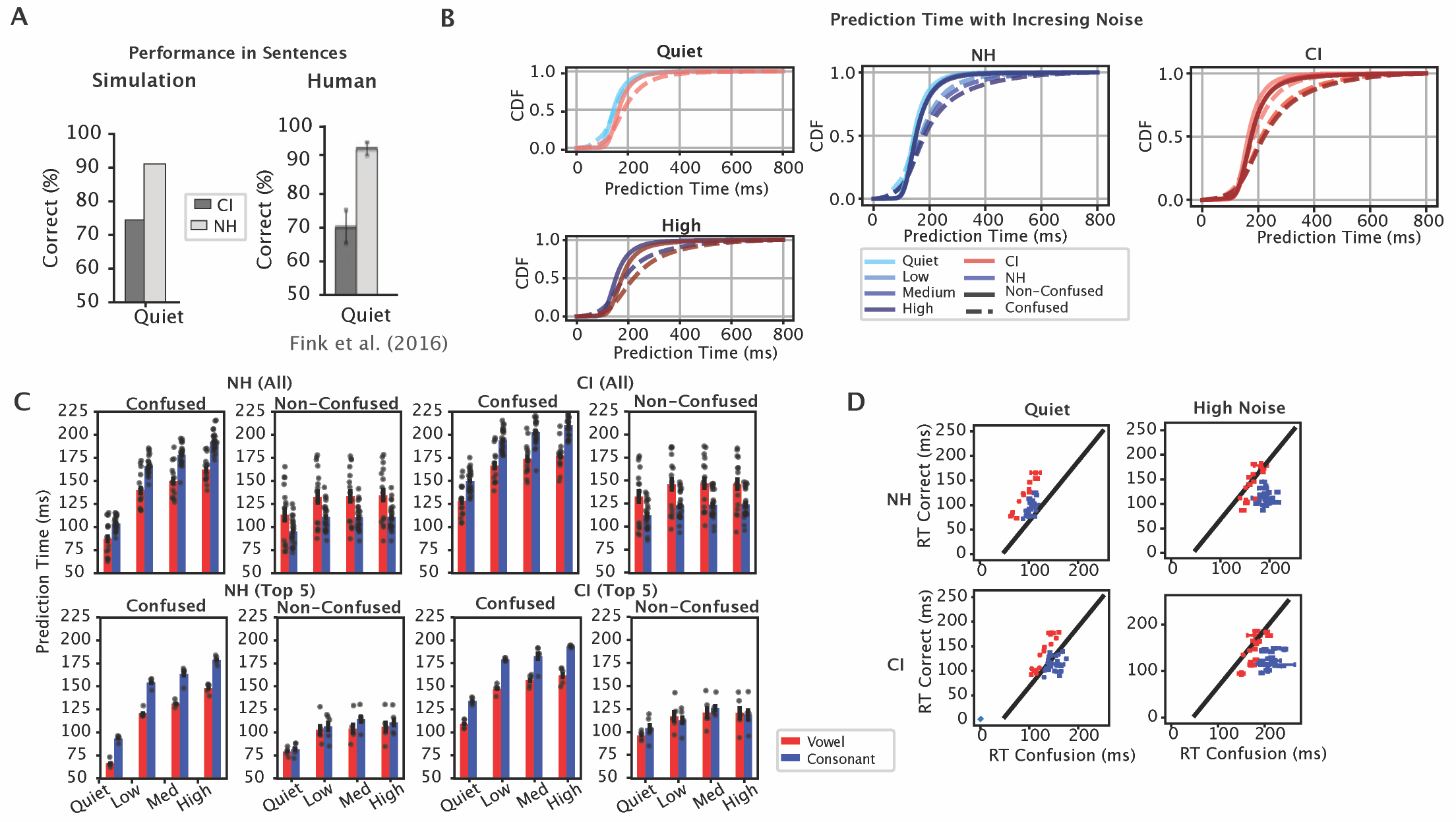}
  %\fbox{\rule[-.5cm]{0cm}{4cm} \rule[-.5cm]{4cm}{0cm}}
  \caption*{Supplemental Figure 2. A. Percent correct performance in spoken sentences of the PhoDe (left) versus humans (right) during phoneme and word recognition task respectively. B. Human non-confusion reaction time data from Finke et al. (2016) \cite{10.1159/000452123} for NH and CI users with increasing noise. C. Reaction time for vowels(red) and consonants(blue) when confused and non-confused plotted against each other for (top) NH and (bottom) CI inputs in (left) quiet and (right) high noise.  D. CDF of reaction time of model on sentences in quiet and high noise levels, showing confused (dashed) versus non-confused (solid) for NH (blue) and CI (red) inputs. Colors increase in darkness with noise intensity.  (Right) all NH responses (left) and all CI responses (right). E. Model reaction time for vowels (red) versus consonants(blue) during confusion or non-confusion for NH(left) and CI(right) inputs. (top) Results for all phonemes in each category. (bottom) Results for the top five shortest reaction times of phonemes per category. }
\end{figure}

\begin{figure}[h]
  \centering
  \includegraphics[width=1\textwidth]{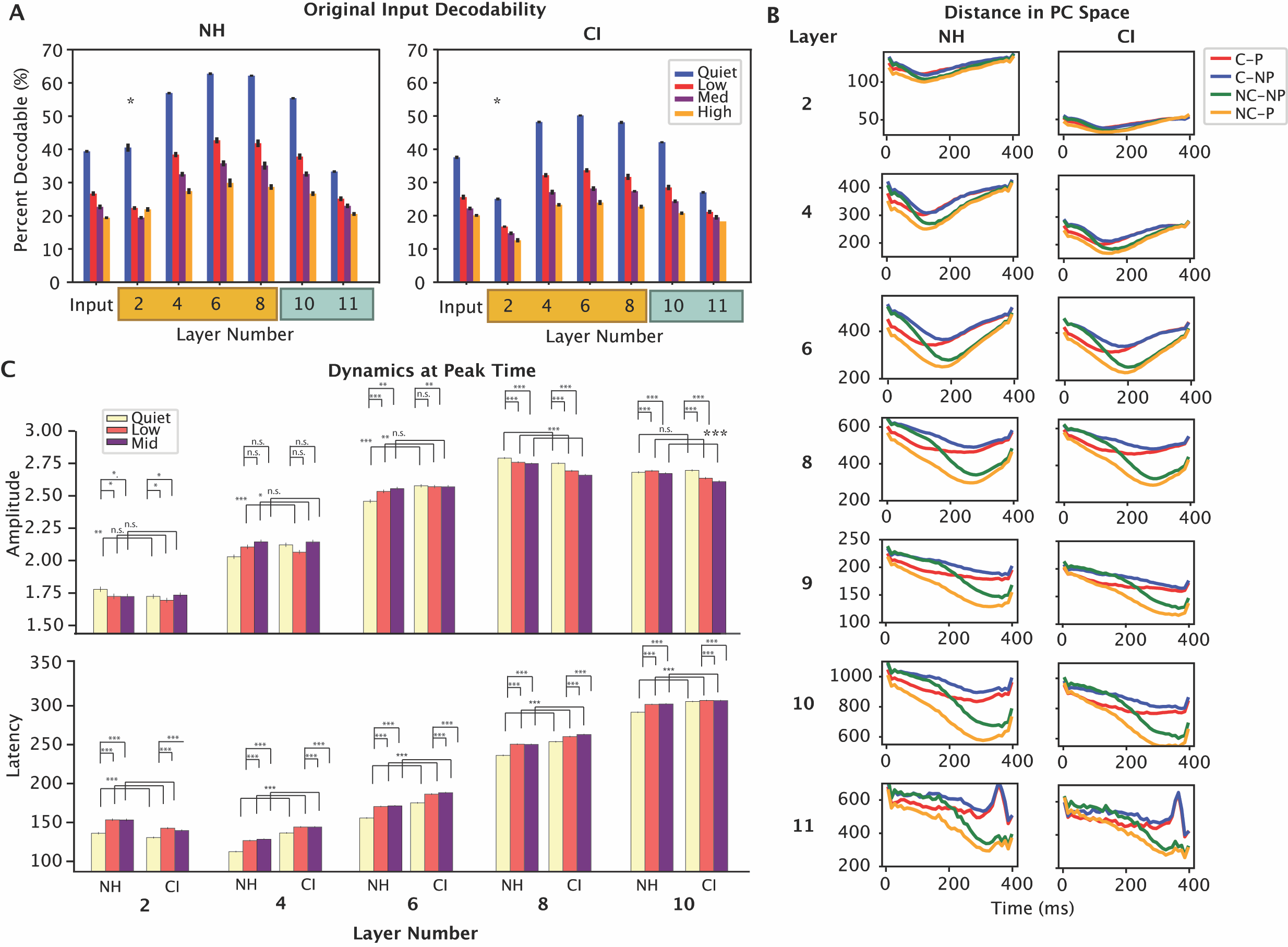}
  %\fbox{\rule[-.5cm]{0cm}{4cm} \rule[-.5cm]{4cm}{0cm}}
  \caption*{Supplemental Figure 3. A. Decodability of inputs per layer of the network compared to the original input. RNN layers (yellow) then linear layers (blue). Decoding performance per layer for NH and CI inputs at increasing  noise levels. Note (*) differences in Layer 2.  B. Distance in PC space between all traces within a category overtime during the utterance window averaged across all phonemes. Layer number increases going down the graphs. NH responses (left) versus CI responses (right) in the NH PC space for C-P(red), C-NP(blue), NC-NP(green), and NC-P(yellow) traces. C. The amplitude (top) and latency (bottom) of the maximal change in response (which the distance between traces is most similar) per layer (left to right) and with increasing levels of noise (quiet - sand, low-peach, mid-purple). Unpaired-test significance shown as p<0.05*,p<0.01**, p<0.001***.}
\end{figure}

\begin{figure}[h]
  \centering
  \includegraphics[width=1\textwidth]{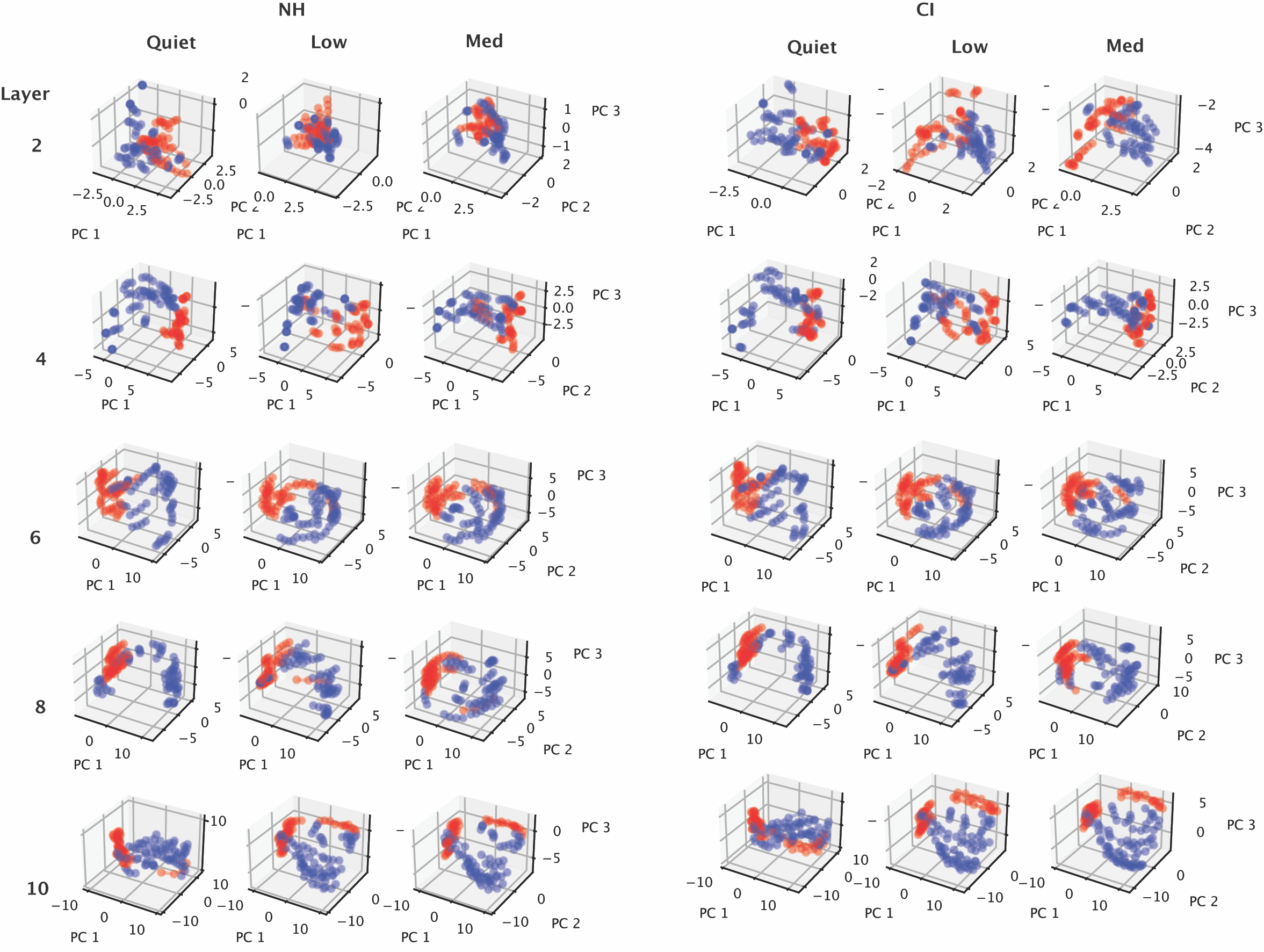}
  %\fbox{\rule[-.5cm]{0cm}{4cm} \rule[-.5cm]{4cm}{0cm}}
  \caption*{Supplemental Figure 4. Projection of dynamics during vowel (\textcolor{red}{red}) and consonants (\textcolor{blue}{blue}) processing per layer during the time point $t_{\text{peak}-1}$ to $t_{\text{peak}+1}$ per layer. Location shown in quiet, low noise, and medium noise going left to right, and from Layer 2 to 10 going down the plot. NH representations (left) compared to CI representations (right) when projected onto the PC space of NH responses only.}
\end{figure}

\subsection{Additional Experiments}
Using a linear decoder shows that after the first LSTM (Layer 2), a significant reduction in decodability of the CI inputs occurs, and noise more negatively affects decodability for CI inputs which is not recovered during deeper layer processing (Supp. Fig. 3). We also observe that in all layers of the network there is a shift in representational space of the phonemes that reduces phoneme separation in NH PC space for CI input (Supp. Fig. 4).
% Optionally include supplemental material (complete proofs, additional experiments and plots) in appendix.
% All such materials \textbf{SHOULD be included in the main submission.}

\FloatBarrier
\section{Supplemental Tables}
\begin{table}[h]
\caption*{Supplemental Table 1: Layer specifications and dimensionality of principal component space with added noise in NH, CI and shared NH+CI conditions}
\label{tab:layerspecs}
\centering
\resizebox{\textwidth}{!}{%
\begin{tabular}{@{}ccccccccccccccccc@{}}
\toprule
\textbf{Layer Num} & \textbf{Layer Type} & \multicolumn{3}{c}{\textbf{Quiet}} & \multicolumn{3}{c}{\textbf{Low}} & \multicolumn{3}{c}{\textbf{Medium}} & \multicolumn{3}{c}{\textbf{High}} & \textbf{\# Outputs} \\
\cmidrule(r){3-5} \cmidrule(lr){6-8} \cmidrule(lr){9-11} \cmidrule(l){12-14} \cmidrule(l){15-15}
 &  & \textbf{NH} & \textbf{CI} & \textbf{NH+CI} & \textbf{NH} & \textbf{CI} & \textbf{NH+CI} & \textbf{NH} & \textbf{CI} & \textbf{NH+CI} & \textbf{NH} & \textbf{CI} & \textbf{NH+CI} &  \\ 
\midrule
0 & Input & 12 & 6 & 18 & 16 & 7 & 23 & 14 & 7 & 21 & 12 & 7 & 19 & 64 \\ 
1 & LSTM1 & \multicolumn{12}{c}{} & 500 \\ 
2 & Batch Norm & 84 & 40 & 124 & 92 & 43 & 135 & 95 & 45 & 140 & 98 & 46 & 144 & 500 \\ 
3 & LSTM2 & \multicolumn{12}{c}{} & 500 \\ 
4 & Batch Norm & 105 & 54 & 159 & 113 & 55 & 168 & 114 & 57 & 171 & 116 & 58 & 174 & 500 \\ 
5 & LSTM3 & \multicolumn{12}{c}{} & 500 \\ 
6 & Batch Norm & 145 & 11 & 156 & 161 & 112 & 273 & 165 & 114 & 279 & 167 & 117 & 284 & 500 \\ 
7 & LSTM4 & \multicolumn{12}{c}{} & 500 \\ 
8 & Batch Norm & 163 & 146 & 309 & 169 & 142 & 311 & 171 & 143 & 314 & 172 & 145 & 317 & 500 \\ 
9 & LSTM5 & \multicolumn{12}{c}{} & 500 \\ 
10 & Batch Norm & 221 & 203 & 424 & 218 & 193 & 411 & 217 & 191 & 408 & 215 & 191 & 406 & 500 \\ 
11 & Fully-Connected & 22 & 20 & 42 & 22 & 20 & 42 & 21 & 19 & 40 & 22 & 20 & 42 & 41 \\ 
\bottomrule
\end{tabular}
}
\end{table}

\begin{table}[h]
\caption*{Supplemental Table 2. Audio augmentation configuration parameters for each noise level}
\label{tab:configparams}
\centering
%\resizebox{\textwidth}{!}{%
\footnotesize
\begin{tabular}{@{}lccc@{}}
\toprule
\textbf{Parameter} & \textbf{Low} & \textbf{Mid} & \textbf{High} \\
\midrule

Background SNR & (10, 15) & (0, 15) & (-10, 15) \\
Pitch Shift & (-2, 2) & (-4, 4) & (-6, 6) \\
Speed Rate & (0.9, 1.1) & (0.7, 1.3) & (0.5, 1.5) \\
Tempo Rate & (0.9, 1.2) & (0.8, 1.4) & (0.7, 1.6) \\
Chorus N & (1, 3) & (1, 4) & (1, 6) \\
Echo N & (1, 3) & (1, 4) & (1, 5) \\
Reverb & (10, 40) & (20, 70) & (30, 100) \\
Low-pass F & (6000, 7500) & (4000, 7000) & (2000, 6000) \\
High-pass F & (100, 500) & (300, 1000) & (500, 2000) \\
Band-pass F & (100, 500) & (200, 1000) & (300, 1500) \\
Band-pass W & (12, 16) & (6, 8) & (3, 5) \\
Band-stop F & (300, 4000) & (300, 2500) & (300, 1500) \\
Band-stop W & (1, 2) & (2, 3) & (3, 5) \\
\bottomrule
\end{tabular}
%}
\end{table}

\end{document}